\documentclass[10pt,twocolumn,letterpaper]{article}

\usepackage{iccv}
\usepackage{times}
\usepackage{epsfig}
\usepackage{graphicx}
\usepackage{amsmath}
\usepackage{mathtools}
\usepackage{amssymb}
\usepackage{mathrsfs} 
\usepackage[standard]{ntheorem}
\usepackage{floatrow}
\usepackage{color}
\usepackage{booktabs}
\usepackage{rotating}

\usepackage[ruled, vlined, linesnumbered]{algorithm2e}
    \SetAlFnt{\small}
    \SetAlCapFnt{\small}
    \SetAlCapNameFnt{\small}

\iccvfinalcopy 

\def\iccvPaperID{1621} 

\ificcvfinal\fi
\begin{document}

\title{Higher-Order Minimum Cost Lifted Multicuts for Motion Segmentation}

\author{Margret Keuper\\
Data and Web Science Group\\
University of Mannheim, Germany\\
{\tt\small keuper@uni-mannheim.de}
}

\maketitle

\begin{abstract}
 Most state-of-the-art motion segmentation algorithms 
 draw their potential from modeling motion differences of local entities such as point trajectories in terms of pairwise potentials in graphical models.
Inference in instances of minimum cost multicut problems defined on such graphs allows to optimize the number of the resulting segments along with the segment assignment. 
 However, pairwise potentials limit the discriminative power of the employed motion models to translational differences. 
 More complex models such as Euclidean or affine transformations call for higher-order potentials and a tractable inference 
 in the resulting higher-order graphical models. In this paper, we (1) introduce a generalization of the minimum cost lifted multicut problem to hypergraphs, 
 and (2) propose a simple primal feasible heuristic that allows for a reasonably efficient inference in instances of higher-order lifted multicut problem instances
 defined on point trajectory hypergraphs for motion segmentation. The resulting motion segmentations improve over the state-of-the-art on the FBMS-59 dataset.
\end{abstract}

\section{Introduction}
Motion segmentation, \ie the task of segmenting all moving objects visible in a video, is a long-standing task in computer vision 
\cite{Bro10c,lezama11,Ochs12,Li2013_12,shiICCV2013}. 
On a low level, it requires an accurate estimation and adequate comparison of the point-wise observable motion. 
At the same time, the number of independently moving objects has to be inferred and all points have to be correctly assigned to a motion segment.
While spectral clustering approaches have been used traditionally in this field \cite{Bro10c,shiICCV2013,Fragkiadaki_2015_CVPR}, 
correlation clustering, also referred to as {\it minimum cost multicut} problem, has recently proven successful on this task \cite{keuper15a}. 

In fact, correlation clustering differs in the objective function from spectral clustering approaches since it does not prefer balanced cuts. 
At the same time, it directly optimizes for the right number of objects, such that there is no need for a separate model selection. 
As spectral clustering, the objective function of the minimum cost multicut problem is defined by edge costs. 
Unlike the spectral clustering scenario, these edge costs can have any real value and thus explicitly model attractive and repulsive terms in the objective function.\footnote{Variants of spectral clustering explicitly incorporating (soft) repulsive terms have been proposed in for example \cite{sc1,sc2}.}
\begin{figure}[t!]
\centering
\includegraphics[width=0.495\linewidth]{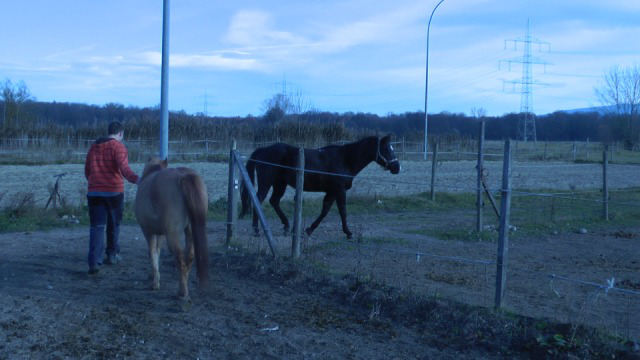} \hfill\phantom{\includegraphics[width=0.495\linewidth]{imageshorses06_0020}}\\[0.3ex]
\includegraphics[width=0.495\linewidth]{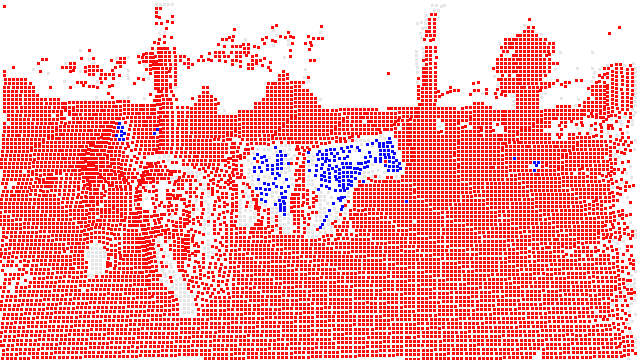}\hfil
\includegraphics[width=0.495\linewidth]{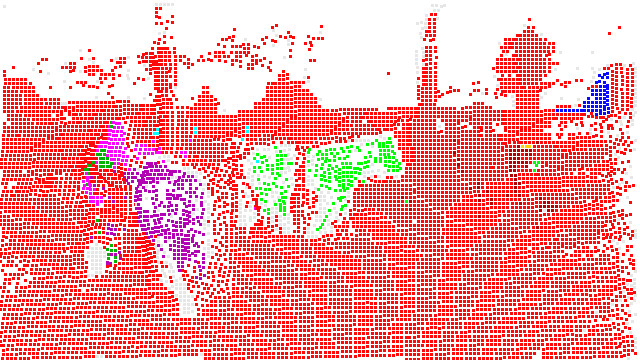}\\[0.3ex]
MCe \cite{keuper15a}\qquad\qquad\qquad\qquad proposed
\caption{Motion segmentation result for a sequence with two consistently moving objects under scaling (the man and the horse next to him). 
The higher-order lifted multicut model allows for a more precise segmentation.}
\label{fig:teaser}
\end{figure}

The formulation of the minimum cost \emph{lifted} multicut model in
\cite{keupericcv} facilitates the definition of such costs between any two
nodes, even if they are not connected in the original graph. This
generalization of the cost function allowed multicut formulations to match the state-of-the-art
in image segmentation and has shown its benefits for multiple target tracking \cite{tang17}. 

One of the key challenges in motion segmentation is to disambiguate between different objects moving according to the same motion model as illustrated in Fig. \ref{fig:teaser}. 
This is comparable to the difficulty in image segmentation when different objects have similar color and texture. 
Lifted Multicuts can dissolve such ambiguities appropriately for image segmentation \cite{keupericcv}. 
In this paper, we show that the same is true for motion segmentation. In Fig. \ref{fig:teaser}, the man and the horse next to him can be segmented from the background 
and from each other although they move similarly.

However, pairwise potentials, such as the lifted multicut model from \cite{keupericcv} allows to define, are very limited in their capacity to describe object motion.
In fact, the Euclidean difference between two local motion descriptors such as optical flow vectors or point trajectories, measures how well the behavior of the two entities 
can be described by a single translational motion model. Depending on the motion recorded in the video, this simple model can yield good performance as shown \eg in \cite{Ochs14,keuper15a}. 
However, when we want to segment objects from videos showing complex motion patterns caused for example by object or camera rotation, scaling motion or zooming, 
more complex motion models are needed. Transformations describing translation, rotation and scaling can be estimated from two motion vectors. 
Thus, for any three points, one can estimate how well their motion can be described by one Euclidean transformation. Edges that describe such motion differences are thus at least of order three. 
Affine motion differences can be described with edges of order four and to assign costs to differences in homographies, the minimum required edge-order is five.

In this paper, we propose the higher-order minimum
cost \emph{lifted}  multicut model. We provide a rigorous definition along with a simple primal feasible heuristic that
allows for practical applicability in computer vision. To the best of our knowledge, 
we are the first to look into this specific problem, which is specially suitable to formulate the motion segmentation objective. 
We demonstrate the benefit of the proposed model on the motion segmentation task using third-order edges and improve over the state-of-the-art on the FBMS-59 \cite{Ochs14} dataset.

\section{Related Work}
Higher-order graph decompositions have been addressed prominently in computer vision since \cite{agarval05}. For example in the subsequent works \cite{agarval06,schollkopf22,chen-2009,Ochs12,zografos14,Elhamifar2009_6,gleichnips16}, spectral approaches 
to the clustering of hypergraphs have been addressed.
Specifically, \cite{Ochs12,zografos14,Elhamifar2009_6} model higher-order motions. Zografos \etal \cite{zografos14} model 3D motions using group invariants and \cite{Elhamifar2009_6} model higher-order motion subspaces. The segmentation is then generated by projecting the resulting hyper-graph onto its primal graph and solving the spectral clustering problem there.
Higher-Order Markov Random Field (MRF) and Conditional Random Field (CRF) models have been proposed for example in \cite{HOMRFFix,HOBP,Schelten2012,HOCRFLiu}.

Our formulation of higher-order minimum cost multicut problems is different from both previous approaches. In contrast to spectral clustering, 
the multicut formulation does not suppose any balancing criterion. Further, we directly infer segmentations from the hyper-graph without any projection onto its primal graph.

In contrast to MRFs, the proposed approach allows higher-order edges to connect vertices globally, violating the Markov property. Further, MRFs and CRFs aim at inferring a node labeling with labels given a priori while multicut approaches aim at inferring an edge labeling yielding an optimal number of segments.   

However, most previous works on minimum cost multicuts in computer vision focus on the definition of problems with pairwise potentials \cite{kappes-2011,keupericcv,keuper15a,tang16}. 
The exception is the model first presented by
\cite{kim-2011,kim-2014}  and extensively studied in \cite{HOsegMC}. Therein, higher-order costs are used for image
segmentation on superpixel graphs with pairwise neighborhood connectivity. In \cite{HOsegMC} an implementation of an ILP solver is proposed for small problem instances.

In contrast, we propose a higher-order {\it lifted} multicut model, which allows the definition of edge costs of any order in the lifted graph as well as in its connectivity defining subgraph. Further, both models differ in that the proposed model allows edges of any order to define the graph connectivity, while in \cite{kim-2011,kim-2014,HOsegMC}, the graph connectivity is defined exclusively by pairwise edges.

Lifted multicuts on simple graphs have been proposed in \cite{keupericcv} along with a primal feasible heuristic to generate solutions. A different heuristic solver for the problem has been proposed in \cite{Beier2016}.
Here, we generalize the solver from \cite{keupericcv} to facilitate the inference in higher-order problems.

In our setup, motion segmentation is cast as a point trajectory grouping problem. In a similar way, it has previously been addressed in \cite{Bro10c,lezama11,Ochs12,Li2013_12,shiICCV2013,Ochs14,Dragon2014,Ji2014,keuper15a}. From sparse motion segmentations, framewise dense segmentations can be computed for example by the variational approaches from \cite{OB11,SO16}.

\section{The Higher-Order Lifted Multicut Problem}
\label{section:model}
Here, we define an optimization problem,
the Higher-Order Minimum Cost Lifted Multicut Problem in analogy to the Lifted Multicut Problem proposed in \cite{keupericcv}.
Its feasible solutions correspond to the decompositions of a hypergraph
and its objective function can assign, for any set $e$ of nodes, a real valued cost to all decompositions for which all nodes $v\in e$ are in the same component.
A component of a hypergraph is any non-empty subgraph that is node-induced and connected by edges of any order.
A decomposition of a hypergraph is any partition $\Pi$ of the node set such that, for every $V' \in \Pi$, the subgraph induced by $V'$ is a connected component of the hypergraph.
An instance of the problem is then defined with respect to an undirected hypergraph $G = (V, F)$ and an additional set of
hyperedges $F'$ connecting sets of nodes that are not necessarily neighbors in $G$.
Practically, we can distinguish between pairwise edges $e \in F_p\cup F'_p $ with $|e|=2$ and higher-order edges $e\in F_h \cup F'_h$ with $|e|>2$. In our experiments, we limit $F$ and $F'$ to edges $e$ of order $|e|=2$ or $3$.
For every edge $e \in E =F\cup F'$, a cost $c_{e} \in \mathbb{R}$ is
assigned to all feasible solutions for which all nodes connected by $e$ are in the same component.

Then, we can define a feasible set $Y_{E} \subseteq \{0,1\}^{E}$ whose elements $y \in Y_{E}$
are 01-labelings of all edges $E =F\cup F'$.
The feasible set is defined such that:
(1) the feasible solutions $y \in Y_{E}$ relate one-to-one to the decompositions of the hypergraph $G$. 
(2) for every edge $e \in E$, $y_{e} = 1$ if and only if all nodes connected by $e$ are in the same component of $G$.
It can be defined rigorously by linear inequality constraints on higher-order edges \cite{kim-2011,kim-2014} \eqref{eq:ho1} and \eqref{eq:ho}, cycle inequalities 
\cite{chopra-1993} \eqref{eq:lmc-cycle} and path and cut constraints \eqref{eq:lmc-path} and \eqref{eq:lmc-cut} \cite{keupericcv}.

\begin{definition}
\label{definition:problem}
For any hypergraph $G = (V, F)$,
any $F' \subseteq \bigcup_k\binom{V}{k} \setminus F$
and any  $c: E := F \cup F' \to \mathbb{R}$,
the 01 linear program written below is an instance of the
\emph{Higher-Order Minimum Cost Lifted Multicut Problem}
w.r.t.~$G$, $E$ and $c$.
\begin{align}
\min_{y \in Y_{E}} 
    & \sum_{e \in E} c_e y_e
    \label{eq:problem}
\end{align}
with $Y_{E} \subseteq \{0,1\}^{E}$ the set of all $y \in \{0,1\}^{E}$ with
\begin{align}
& \hspace{-1ex}\forall e, e_h\in E\ |\ e\subset e_h: y_{e_h}\leq y_{e} \ \label{eq:ho1}\\
& \hspace{-1ex} \forall {e_h} \in E\ |\ \bar{G}=({e_h},e\in E|e\subset {e_h})\hspace{2ex} \text{is connected}:\ \nonumber\\
& \hspace{1ex} (1-y_{{e_h}})\leq \sum_{e\in E|e\subset {e_h}} (1-y_{e}) 
\label{eq:ho}\\
& \hspace{-1ex} \forall C \in \textnormal{cycles}(G)\ \forall e \in C:\ 
    (1-y_e) \leq \hspace{-2ex} \sum_{e' \in C \setminus \{e\}} \hspace{-2ex} (1-y_{e'})
    \label{eq:lmc-cycle}\\
  & \hspace{-1ex} \forall e\in F, \forall vw \subset e\forall P \in vw\textnormal{-paths}(G):\ \nonumber\\
  &   \hspace{1ex}(1-y_{e}) \leq \sum_{e' \in P}(1- y_e')
    \label{eq:lmc-path}\\
& \hspace{-1ex} \forall e\in F, \forall vw \subset e \forall C \in vw\textnormal{-cuts}(G):\ \nonumber\\
   & \hspace{1ex} y_{e} \leq \sum_{e' \in C} (y_e') 
    \label{eq:lmc-cut}
\end{align}
\end{definition}

Note that, since higher-order edges $e\in F$ are connectivity defining, these conditions need to hold on cycles, $vw$-paths (any path in $G$, connecting $v$ and $w$) and $vw$-cuts (any cut in $G$ assigning $v$ and $w$ to distict components) of any order.
In the higher-order model from \cite{kim-2011}, it is sufficient to ensure the cycle inequalities \eqref{eq:lmc-cycle} for cycles on pairwise edges.
For the motion segmentation application, it is reasonable to allow 
higher-order edges to define the graph connectivity since most motion models can only be estimated on at least two nodes. If, in this case, for a pair of 
nodes $v$ and $w$ all higher-order edges with $\{v,w\}\subseteq e$ have label $y_e =0$, 
there was no motion model that supports $v$ and $w$ belonging to the same object. Consequently, any grouping of a subset of $e$, for which the motion model can not be evaluated, to the same component, is uninformed and should thus be avoided. However, if the connectivity is defined from the projection of $e$ onto a primal graph, this connection is possible and cost-neutral. 

The definition of lifted edges allows to ensure a consistent motion segment connectivity. For example background motion can often only be correctly 
estimated from points with large spatial distance. In this case, the constraints \eqref{eq:lmc-cut} ensure an existing path between these background points. 
At the same time, defining the connectivity on sets of points in a spatial neighborhood allows to cut apart distinct but consistently moving objects.

In the following, we describe a primal feasible heuristic that allows to solve instances of the Higher-Order Lifted Multicut Problem in practice.

\section{Higher-Order Lifted Kernighan-Lin Algorithm}

The Kernighan-Lin~\cite{kernighan-1970} heuristic is known to work well for the balanced set partitioning problem. The original heuristic has been modified in \cite{openGM} to disregard the size of the resulting partitions as well as to optimize for the number of components. In \cite{keupericcv}, it has been further adapted to work with sparse and lifted graphs; an additional joining move has been introduced for a more stable behavior with different initializations.
Here, we further extend the algorithm to work with \emph{higher-order} lifted graphs. While the outer loop of the algorithm, running over all pairs of neighboring components is only marginally affected, the major change concerns the inner iterations, where the gain of moving vertices is computed and successively updated. 
The algorithm takes as input an instance of the higher-order lifted multicut problem and an initial decomposition of $G$ and outputs a decomposition of $G$ whose higher-order lifted multicut has an objective value lower than or equal to that of the initial decomposition. 
As the basic KLj-Algorithm \cite{keupericcv}, it maintains, throughout its execution, a decomposition of $G$
which is encoded as a graph $\mathcal{G} = (\mathcal{V}, \mathcal{E})$. The nodes $a \in \mathcal{V}$ of $\mathcal{G}$ are components of $G$ and its edges $ab \in \mathcal{E}$ connect all components $a$ and $b$ of $G$ which are connected in $G$ by an edge of arbitrary order.

\begin{algorithm}[t]
	\DontPrintSemicolon
	\KwData{weighted undirected lifted higher-order graph $G=(V,F \cup F',c)$, starting 01-edge labeling $y^{|F \cup F'|}$}
	\KwResult{01-edge labeling $y$}
	\BlankLine
        $t \leftarrow 1$
	\While{$t < max\_iter$ {\bf and not} $y^t = y^{t-1}$}
	{
          \ForEach{$(a,b) \in$ adjacent\_partitions($y^{t-1}$)}
          {
            
            $y^t \leftarrow$ higher\_order\_update\_bipartition($G, a, b$)\;
          }
          
          \ForEach{$a \in$ partitions($y^{t}$)}
          {

              $y^t \leftarrow$ higher\_order\_update\_bipartition($G, a, \emptyset$)\;
            
          }
          
	}
	\caption{Kernighan-Lin Algorithm. The function higher\_order\_update\_bipartition is given in Algorithm~\ref{alg:KL_inner}.}
	\label{alg:KL_outer}
\end{algorithm}
\begin{algorithm}[t]
	\DontPrintSemicolon
	\KwData{weighted undirected lifted higher-order graph $G=(V,F \cup F',c)$, a pair of partitions $A$ and $B$}
	\KwResult{01 - edge labeling $y$}
	\BlankLine
	$D^{A\cup B}$ = compute\_differences($F,F', A, B$)\;
        $BDY \leftarrow$ compute\_boundary\_nodes($F, A, B$)\;
	$\Delta_{join} \leftarrow$ compute\_gain\_from\_joining($F,F', A, B$)\;
	$S^{|A \cup B|}$ = 0\tcp*[r]{cumulative gain}
        $M = [.]$ \tcp*[r]{empty move vector};
	\For{$i \leftarrow 1$ \KwTo $|BDY|$}
	{
		$v^* \leftarrow \mathrm{argmax}_{v}$ ($D_{(A\cup B)\cap BDY}$)\;
                $M$.push\_back($v^*$)\;
		\ForEach{e $\in$ G.edges($v*$)}
		{
                  \If{same\_partition(G.nodes($e$)$\setminus v^*$)}
                  {

                    \ForEach{w $\in$ G.nodes($e$)$\setminus v^*$}
                    {
                      \If{\text{same\_partition}($v^*,w$)  {\bf and} $|\text{G.nodes}(w) \setminus v^*|$ = 1 }
                      {
                        $D_w \leftarrow D_w - 2c_{e}$\;
                      }
                      \ElseIf{same\_partition($v^*,w$)  {\bf and} $|\text{G.nodes}(w) \setminus v^*| >$  1 }
                      {
                        $D_w \leftarrow D_w - c_{e}$\;
                      }
                      \ElseIf{ {\bf not} same\_partition($v^*,w$)  {\bf and} $|\text{G.nodes}(w) \setminus v^*|$= 1 }
                      {
                        $D_w \leftarrow D_w + 2c_{e}$\;
                      }
                      \ElseIf{{\bf not} same\_partition($v^*,w$)  {\bf and} $|\text{G.nodes}(w) \setminus v^*| >$  1 }
                      {
                        $D_w \leftarrow D_w + c_{e}$\;
                      }
                    }
                  }
                  \Else{
                    \ForEach{w $\in$ G.nodes($e$)$\setminus v^*$}
                    {
                      \If{same\_partition(G.nodes($e$)$\setminus v^*,w$)}{
                        \If{same\_partition($v^*,w$) }
                           {
                             $D_w \leftarrow D_w + c_{e}$\;
                           }
                     
                         \ElseIf{{\bf not} same\_partition($v^*,w$) }
                           {
                             $D_w \leftarrow D_w - c_{e}$\;
                           } 
                               
                      }
                    }
                  }  
                }
                $S_i = S_{i-1} + D_{v^*}$\;
                $BDY \leftarrow$ update\_boundary($v^*, F, A, B$)\;
	}

	$k \leftarrow \mathrm{argmax}_i S_i$\tcp*[r]{best number of moves}
	\If{$\Delta_{join} > S_{k}$ {\bf and} $\Delta_{join} > 0$}
	{
		join\_partitions($y, A, B$);
	}
	\ElseIf{$S_{k} > 0$}
	{
		move\_nodes($y, A, B, k$)\;
	}
	\caption{Function higher\_order\_update\_bipartition}
	\label{alg:KL_inner}
\end{algorithm}

Algorithm~\ref{alg:KL_outer}
starts from an initial decomposition provided as input.
As KLj \cite{keupericcv}, in each iteration, it tries to improve the current decomposition by one of these three transformations:
(1)~moving a set of nodes between two neighboring components,
(2)~moving a connected set of nodes from one component to a new component,
(3)~joining two neighboring components.
The main operation of
Alg.~\ref{alg:KL_outer} is called ``higher\_order\_update\_bipartition''.
Its input is the current decomposition and a pair $ab \in \mathcal{E}$ of neighboring components of $G$. The neighborhood in $G$ is defined with respect to edges $e \in F$ of any order. It evaluates transformations (1) and (3) for $ab$. Transformation (2) is assessed by executing ``higher\_order\_update\_bipartition'' for each component and $\emptyset$.

In the operation ``higher\_order\_update\_bipartition'' 
a sequence $M$ of elementary transformations of the components $a$ and $b$ is constructed greedily such that at every consecutive move-operation increases the cumulative gain $S$ maximally (or decreases it minimally) while preserving a feasible solution. 
Therefore, the operation ``compute\_differences'' computes, at the beginning of each execution of ``higher\_order\_update\_bipartition'', for every element $w \in a \cup b$ the difference $D_w$ in the objective function when $w$ is moved between $a$ and $b$. Then, an element on the current boundary between $a$ and $b$ with maximal difference is added to the sequence $M$. The differences $D$ are updated according to Alg. \ref{alg:KL_inner}, ll. 9-26. 

If the objective value can be decreased by executing either the first $k \in \mathbb{N}_0$ elementary transformations or by joining the components $a$ and $b$ the optimal of these two operations is carried out.

While components are defined with respect to the graph $G = (V,F)$,
differences in objective value are computed with respect to the graph $G' = (V, E)$ with $E=F \cup F'$.

As in KLj \cite{keupericcv}, the number of outer iterations of 
Alg.~\ref{alg:KL_outer} 
is not bounded by a polynomial and we cannot give any guarantee for convergence. However, in practice, the algorithm converged in less than 50 iterations for the experiments described in 
Sec.~\ref{section:experiments}.
\section{Graph Construction}
\subsection{Point Trajectories}
Point trajectories are spatio-temporal curves that describe the trajectory of a single object point in the image plane. 
They build the basis for many motion segmentation methods such as \cite{Bro10c,Fragkiadaki_2015_CVPR,Fragkiadaki_videosegmentation,Ochs14,keuper15a}. 
Here, we use the method from \cite{Bro10c} to generate dense long-term point trajectories from precomputed optical flow. 
While we are aware that more recent optical flow algorithms allow for better motion segmentations \cite{flownet}, 
we build our trajectories from large displacement optical flow \cite{ldof} to allow for a direct comparison to previous work.
For a video of length $N$, \cite{Bro10c} yields $n$ point trajectories $p_i$ with the maximum length $N$, where $n$ depends on the desired sampling rate. 
Due to occlusions and mistakes in the optical flow estimation, most trajectories are significantly shorter than $N$, 
and some trajectories start after frame 1 to ensure even point sampling throughout the sequence. 
\subsection{Higher-Order Motion Models}
Although this is not sufficient to accurately describe object motion in a 3D environment recorded with a possibly moving camera,
we restrict ourselves to edge potentials of order two and three for practical reasons. 
This allows to measure the difference of point motions according to Euclidean motion models, \ie from the group of transformations describing translation, rotation and scaling in the 2D plane. 
This is a subset of the group of similarity transformations in the 2D plane, where reflections are excluded.

We further argue that in any case, the easiest model that can explain the motion of a set of points with a single transformation should be used. 
If two points are moving according to the same translational motion model, we can assume that they  belong to the same object without looking at further points around them.
Only if their motion is different according to a purely translational model, looking at more complex motion models adds information. 
This results in a {\it motion-adaptive graph construction} strategy.
\paragraph{Motion-Adaptive Graph Construction}
We propose to construct the higher-order graph $G$ depending on the pairwise costs computed from motion differences. The algorithm is described in Alg. \ref{alg:graph}. For any pair of trajectories, 
we compute their cost of belonging to the same translational motion model. Only if this cost is repulsive, 
we look at all further points to compute for every three-tuple the cost of belonging to the same motion model for translation, rotation and scaling. 
The respective third-order edges are inserted along with their costs.

This strategy allows to integrate second and third order potential without losing model capacity. 
Further, compared to generating the full graph with higher-order potentials, it yields a huge complexity reduction in practice.

\begin{algorithm}[t]
	\DontPrintSemicolon
	\KwData{set of point trajectories $V$ with $p_k\in V$ with $k\in \{1 \dots n\}$}
	\KwResult{weighted undirected higher-order graph $G=(V,E)$, cost vector $c$}
	\BlankLine
	$G\leftarrow (V,E=\emptyset)$\;
	$c=[.]$\;
	\ForEach{$p_i$ and $p_j \in V$}
          {
          $c_{ij}\leftarrow$ compute\_translational\_motion\_cost($p_i, p_j$)\;
          $E\leftarrow E\cup(p_i, p_j)$\tcp*[r]{insert pairwise edge}
          \If{$c<0$}
          {
	    $c$.push\_back($c_{ij}$)\;
	  }
	  \Else
	  {
	     $c$.push\_back($0$)\;
	    \ForEach{$p_k \in V\ \{p_i,p_j\}$}
	      {
		$c_{ijk}\leftarrow$ compute\_higher\_order\_motion\_cost($p_i, p_j, p_k$)\; 
		$E\leftarrow E\cup (p_i, p_j, p_k)$\tcp*[l]{insert higher-order edge} 
	        $c$.push\_back($c_{ijk}$)\;
	       }
	   }
	 }
	\caption{Motion Adaptive Graph Construction.}
	\label{alg:graph}
\end{algorithm}
\paragraph{Lifted Graph Construction}
To construct higher-order lifted graphs $G' = (V, F \cup F')$, we precompute for every trajectory the set of its 12 spatially nearest neighbors $\mathcal{N}$. The edge set $F$ is the subset of the full edge set $E$, computed according to Alg. \ref{alg:graph}, that contains exactly all pairwise edges $e_{ij}\in E$ for which at least one of the following three conditions holds: (1) $p_i \in \mathcal{N}(p_j)$, (2) $p_j \in \mathcal{N}(p_i)$ (3) the maximum spatial distance between $p_i$ and $p_j$ is below 40 pixels.

\paragraph{Second Order Costs}
Second order costs are computed from pairwise differences on point trajectories. We compute such differences only for trajectories which have at least two frames in common.  
Since it has proven successful in previous work \cite{keuper15a}, we compute such differences based on motion, color and spatial distance cues. As suggested by \cite{Ochs14}, we define the pairwise motion difference of two trajectories at time $t$ as
\begin{equation}
d_t^{\text{m}}(p_i,p_j)=\frac{\|\partial_tp_i - \partial_tp_j\|}{\sigma_t}.
\label{eq:motiondist}
\end{equation} 
Here, $\partial_tp_i$ and $\partial_tp_j$ are the partial derivatives of $p_i$ and $p_j$ with respect to the time dimension
and $\sigma_t$ is the variation of the optical flow as defined in \cite{Ochs14}.
The motion distance of two trajectories is defined by the maximum over time

\begin{equation}
d^{\text{m}}(p_i,p_j) = \max_t d^{\text{m}}_t(p_i,p_j).
\label{eq:motiondistmax}
\end{equation}

As proposed in \cite{keuper15a} color and spatial distances $d^{\text{c}}$  and $d^{\text{s}}$ are computed as average distances over the common lifetime of two trajectories. These three cues are combined non-linearly to compute the costs

\begin{align}
c_{ij}&=&-\max\left(\right.&{\bar \theta_0}                  &+\,&\theta_1d^{\text{m}}(p_i,p_j)\,+\\
     & &                  &\theta_2       d^{\text{s}}(p_i,p_j)&+\,&\theta_3d^{\text{c}}(p_i,p_j)\quad, \nonumber\\
     & &                  &\theta_0                         &+\,&\theta_1d^{\text{m}}(p_i,p_j)\quad\left.\right) \nonumber
\label{eq:distmax}
\end{align}
with weights and intercept values $\theta$ as proposed in \cite{keuper15a}.

\paragraph{Third Order Costs}
We compute third order motion differences as proposed in \cite{Ochs12}. For any two trajectories $p_i$ and $p_j$ coexisting from time $t$ to $t'$, we estimate the Euclidean motion model $\mathcal{T}_{ij}(t)$, consisting of rotation $R_\alpha$, translation $v:=(v_1,v_2)^\top$  and scaling $s$ as

\begin{align}
\alpha &=&\text{arccos}\left(\frac{(p_i(t')-p_j(t'))^\top(p_i(t)-p_j(t))}{\|p_i(t')-p_j(t')\|\cdot\|p_i(t)-p_j(t)\|}\right)\\
s&=&\frac{\|p_i(t')-p_j(t')\|}{\|p_i(t)-p_j(t)\|} \nonumber\\
v&=&\frac{1}{2}\left(p_i(t') + p_j(t') - sR_\alpha(p_i(t) + p_j(t))\right) \nonumber
\end{align}

The distance to any third trajectory $p_k$ existing from $t$ to $t'$ can then be measured by $d^t_{ij}(p_k) = \|\mathcal{T}_{ij}(t)p_k(t) - p_k(t')\|$. For numerical reasons,  $d^t_{ij}(p_k)$ is normalized by 
\begin{align}
\gamma^t_{ij}=\frac{1}{\sigma_t}\left(\frac{1}{2}\left(\frac{\|p_i(t)-p_j(t)\|}{\|p_i(t)-p_k(t)\|} +\frac{\|p_i(t)-p_j(t)\|}{\|p_j(t)-p_k(t)\|} \right)\right)^{\frac{1}{4}}, 
\end{align}
with $\sigma_t$ being the optical flow variation as in \eqref{eq:motiondist}. 

To render distances symmetric, \cite{Ochs12} propose to consider the maximum $d^t_{\text{max}}(i,j,k) = \text{max}(\gamma^t_{ij}d^t_{ij}(p_k), \gamma^t_{ik}d^t_{ik}(p_j),\gamma^t_{jk}d^t_{jk}(p_i))$, which yields an over-estimation of the true distance. While this is unproblematic in a spectral clustering scenario, where distances are used to define positive point affinities, it can lead to problems in the multicut approach. Over-estimated distances lead to under-estimated join probabilities and thus eventually to switching the sign of the cost function towards repulsive terms. To avoid this effect, we compute both $d^t_{\text{max}}(i,j,k)$ and, analogously, $d^t_{\text{min}}(i,j,k)$.
For both, we compute the maximum motion distance over the common lifetime of $p_i$, $p_j$ and $p_k$ as $d_{\text{max}}(i,j,k) = \text{max}_td^t_{\text{max}}(i,j,k)$ and $d_{\text{min}}(i,j,k) = \text{max}_td^t_{\text{min}}(i,j,k)$. We evaluate the costs $c(d_{\text{max}}(i,j,k))$ and $c(d_{\text{min}}(i,j,k))$ for both distances as $c(d) = \theta_0 +\theta_1d$ and compute the final edge costs 

\begin{align}
c_{ijk} = \begin{cases*}
c(d_{\text{min}}(i,j,k))& if\qquad $c(d_{\text{max}}(i,j,k)) > 0$\\
c(d_{\text{max}}(i,j,k))& if\qquad $c(d_{\text{min}}(i,j,k)) < 0$\\
0&otherwise.
\end{cases*}
\label{eq:thirdordercosts}
\end{align}
Thus, we make sure not to set any costs for edges whose underlying motion is controversial. Here, we set $\theta_0 = -1$ and $\theta_1 = 0.08$ manually.
\vspace{-0.2cm}
\paragraph{Implementation Details}
In practice, we insert pairwise edges $e_{ij}$ in $G$ and $G'$ only if the spatial distance between $p_i$ and $p_j$ is below 100 pixel. 
This is in analogy to \cite{keuper15a} and due to the fact that for nearby points, the approximation of the true motion by a simplified model is 
usually better than for points at a large distance. Also, since the number of pairwise edges increases quadratically with the maximal spatial distance, 
this heuristic decreases the computational load significantly. For the same reason, we introduce an edge sampling strategy for third order edges. 
For every three-tuple of points, we compute the maximum pairwise distance $d$. From all three-tuples with $20<d<300$, we randomly sample $\frac{100}{d^2} \%$, while we insert all edges $e_{ijk}$ with $d\leq 20$. This also prevents from a too strong imbalance of long range edges over short range edges. 
\section{Experiments }
\label{section:experiments} 
%
We evaluate the proposed higher-order lifted multicut model on the motion segmentation benchmark FBMS-59 \cite{Ochs14},
which is an extended version of the BMS-26 benchmark from Brox-Malik \cite{Bro10c}. It contains 59 sequences of varying length 
(from 19 to 800 frames) and diverse content and motion. It provides manual annotations for all moving objects in the videos for 
every 20th frame. To allow for training, the dataset has been split into two subsets of 29 and 30 sequences for training and testing, respectively. 
While we agree that training all model parameters is highly desirable, we did not do so. This is due to the fact that
(1) neither of the state-of-the-art methods \cite{Ochs12,Ochs14,keuper15a} is training-based and 
(2), the training set, with 29 sparsely annotated sequences, is rather small. 
Thus, to avoid confusion, we hence denote the training split by {\it Set A} and the test split by {\it Set B}. 

\begin{table}[tb]
\centering
\small
\scalebox{1}{
\begin{tabular*}{\linewidth}{@{\extracolsep{\stretch{1}}}lcccc@{\extracolsep{\stretch{1}}}}
\toprule

 \textbf{Set A} (29 sequences)   &  \small{P}  & \small{R}  &\small{F}  & \small{O}\\                            
\cmidrule(lr){2-5}
 \small{SC \cite{Ochs14}}& \small{85.10\%} & \small{62.40\%} & \small{72.0\%}  & \small{17/65} \\
\small{Higher-Order SC \cite{Ochs12}}& \small{81.55\%} & \small{59.33\%} & \small{68.68\%}  & \small{16/65} \\
\small{MC \cite{keuper15a}} &  \small{84.94\%} & \small{71.22\%} & \small{77.48\%}  & \small{23/65} \\
\small{HO MC (ours)} &  \small{83.51\%} & \small{75.54\%} & \small{\bf 79.33\%}  & \small{\bf 28/65} \\
\midrule

    \textbf{Set B} (30 sequences)                  &  \small{P}  & \small{R}  &\small{F}  & \small{O}\\

\cmidrule(lr){2-5}
 \small{SC \cite{Ochs14}}&   \small{79.61\%} & \small{60.91\%} & \small{69.02\%} & \small{24/69}  \\
\small{Higher-Order SC \cite{Ochs12}}&    \small{82.11\%} & \small{64.67\%} & \small{72.35\%} & \small{\bf 27/69}\\
\small{MC \cite{keuper15a}} &  \small{82.87\%} & \small{69.89\%} & \small{\bf 75.83\%} & \small{\bf 27/69} \\
\small{HO MC (ours)} &    \small{83.62\%} & \small{68.83\%} & \small{75.51\%} & \small{\bf 27/69} \\
\bottomrule
\end{tabular*}
}
\caption{\label{tab:hoeval} Segmentation results on the FBMS-59 dataset on Set A (top) and Set B (bottom). We report \textbf{P}: average precision, \textbf{R}: average recall, \textbf{F}: F-measure and \textbf{O}: extracted objects with $\text{F}\geq 75\%$.  All results are computed for sparse trajectory sampling at 8 pixel distance. Our result {\bf HO MC} is computed on the non-lifted purely higher-order model to allow for a direct comparison to the listed competing methods.}
\end{table}

\paragraph{Evaluation}
To assess the capacity of our model components, we first evaluate a purely higher-order non-lifted version of our model. 
In this model, all pairwise costs are removed and all edges are connectivity defining. 
We compare this simple model to \cite{Ochs14,Ochs12} and the purely motion-based version of \cite{keuper15a}. 
While \cite{Ochs14} and \cite{keuper15a} only consider translational motion, 
the affinities in \cite{Ochs12} are defined most similarly to our higher-order costs. 
As the proposed approach, \cite{keuper15a} formulate a multicut problem while \cite{Ochs14,Ochs12} follow a spectral clustering approach. 
The results are given in Tab.~\ref{tab:hoeval} in terms of precision, recall, f-measure and the number of extracted objects. 
Precision and recall are not directly comparable but they can serve as cues for under- or over-segmentation. The f-measure is an aggregate of both. 
From Tab.~\ref{tab:hoeval}, we can observe that our higher-order lifted multicut model outperforms the higher-order spectral clustering method from \cite{Ochs12} by about 10\% on Set A and 3.5\% on Set B.
While there is a clear improvement on both sets, the imbalance is remarkable. A similarly remarkable imbalance can be observed when comparing the performance of the two spectral clustering methods \cite{Ochs14} and \cite{Ochs12}. The higher-order model \cite{Ochs12} yields a lower f-measure on Set A compared to \cite{Ochs14}. Yet it outperforms \cite{Ochs14} on Set B by about 3\%. This indicates that the motion statistics in both splits are significantly different.

When we compare our higher-order model to the pairwise minimum cost multicut model from \cite{keuper15a}, we can observe an improvement on Set A. On Set B, both model perform almost equally in terms of f-measure.

\begin{table}[tb]
\centering
\small
\scalebox{1}{
\begin{tabular*}{\linewidth}{@{\extracolsep{\stretch{1}}}lcccc@{\extracolsep{\stretch{1}}}}
\toprule
 \textbf{Set A} (29 sequences)   &  \small{P}  & \small{R}  &\small{F}  & \small{O}\\                            
\cmidrule(lr){2-5}
\small{MCe \cite{keuper15a}} &  \small{86.73\%} & \small{73.08\%} & \small{79.32\%}  & \small{31/65} \\
\small{HOPMC} &  \small{86.45\%} & \small{76.28\%} & \small{81.05\%}  & \small{31/65} \\
\small{AOMC} &  \small{83.91\%} & \small{76.44\%} & \small{80.00\%}  & \small{\bf 33/65} \\
\cmidrule(lr){2-5}
\small{Lifted HOPMC} &  \small{88.19\%} & \small{77.10\%} & \small{\bf 82.27\%}  & \small{31/65} \\
\small{Lifted AOMC} &  \small{86.83\%} & \small{77.79\%} & \small{82.06\%}  & \small{32/65} \\
\midrule

    \textbf{Set B} (30 sequences)                  &  \small{P}  & \small{R}  &\small{F}  & \small{O}\\

\cmidrule(lr){2-5}
\cmidrule(lr){2-5}
\small{MCe \cite{keuper15a}} &  \small{87.88\%} & \small{67.7\%} & \small{76.48\%} & \small{25/69} \\
\small{HOPMC} &    \small{85.83\%} & \small{68.74\%} & \small{76.34\%} & \small{\bf 26/69} \\
\small{AOMC} &    \small{84.19\%} & \small{72.64\%} & \small{77.99\%} & \small{\bf 26/69} \\
\cmidrule(lr){2-5}
\small{Lifted HOPMC} &  \small{85.88\%} & \small{69.24\%} & \small{76.68\%}  & \small{\bf 26/69} \\
\small{Lifted AOMC} &    \small{87.77\%} & \small{71.96\%} & \small{\bf 79.08\%} & \small{25/69} \\
\bottomrule
\end{tabular*}
}


\caption{\label{tab:msevaluation} Segmentation Results on FBMS-59 on Set A (top) and Set B (bottom). 
We report \textbf{P}: average precision, \textbf{R}: average recall, \textbf{F}: F-measure and \textbf{O}: extracted objects with $\text{F}\geq 75\%$. 
All results are computed for sparse trajectory sampling at 8 pixel distance.}
\end{table}
In Tab.~\ref{tab:msevaluation}, we show the evaluation of our higher-order multicut model with the motion-adaptive order, denoted AOMC (compare Alg.~\ref{alg:graph}). 
This model has access to similar pairwise cues as the motion 
and color-based version from \cite{keuper15a}, denoted MCe. It has as well access to the higher-order motion cues from Eq.~\eqref{eq:thirdordercosts}.

As a sanity check for the motion-adaptive graph construction, we generate graphs that simply contain all pairwise costs $c_{ij}$ 
as well as all third order edges with costs $c_{ijk}$ without any adaptation w.r.t. the costs. We denote this additive model by HOPMC (higher-order + pairwise multicut). 
On Set A, both variants improve over MCe~\cite{keuper15a} while on Set B, HOPMC performs similarly as MCe~\cite{keuper15a} and is outperformed by AOMC by about 1.5\% in f-measure.

Lifted versions of both types of proposed problems (HOPMC and AOMC) yield a further improvement on both sets w.r.t. the respective non-lifted model.
However, on Set B, the segmentation quality of Lifted HOPMC does not significantly outperform the one of MCe. In contrast, the proposed Lifted AOMC 
consistently outperforms all competing methods and baselines.

\begin{figure*}[ht!]
\centering
\scalebox{1}{
\centering
\begin{tabular}{@{}l@{\,}l@{\,}l@{\,}l@{\,}l@{\,}l@{}}
\includegraphics[height=1.9cm]{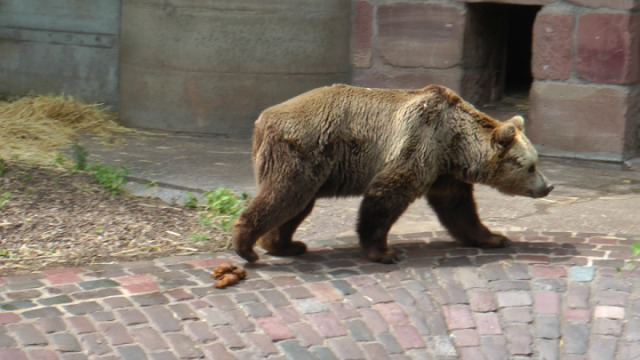}&
\includegraphics[height=1.9cm]{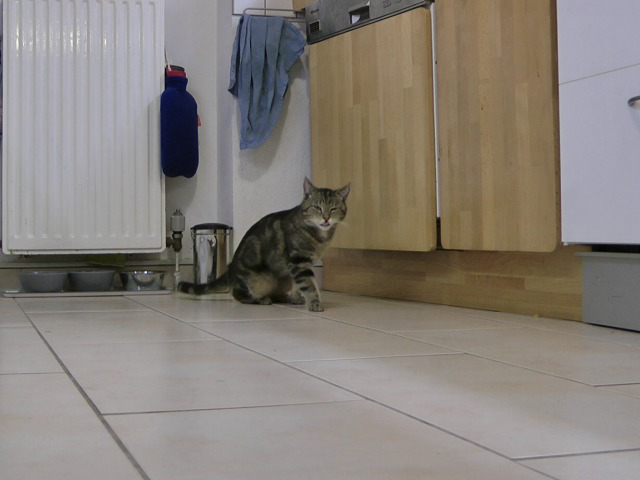}&
\includegraphics[height=1.9cm]{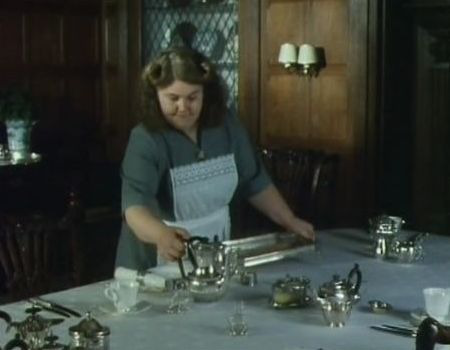}&
\includegraphics[height=1.9cm]{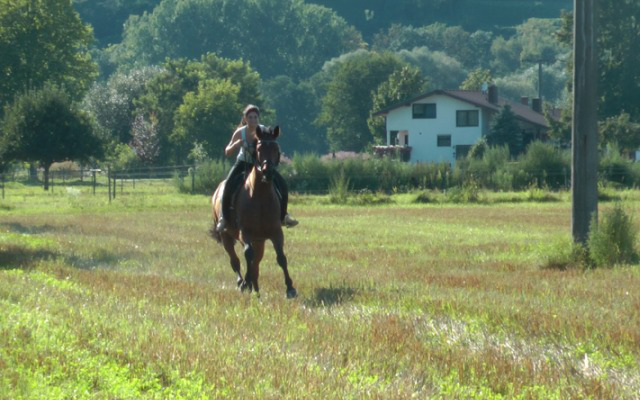}&
\includegraphics[height=1.9cm]{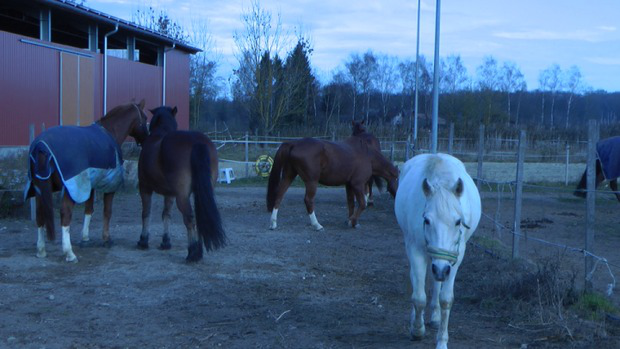}&
\includegraphics[height=1.9cm]{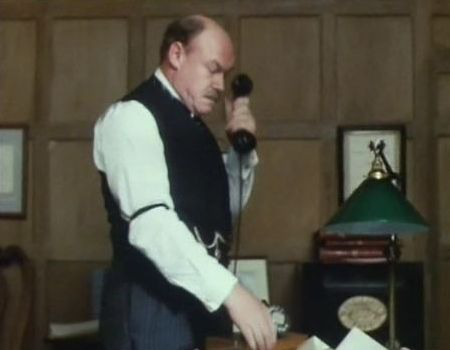}
\\
\includegraphics[height=1.9cm]{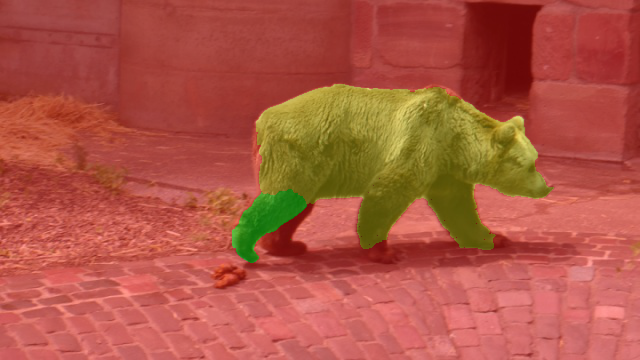}&
\includegraphics[height=1.9cm]{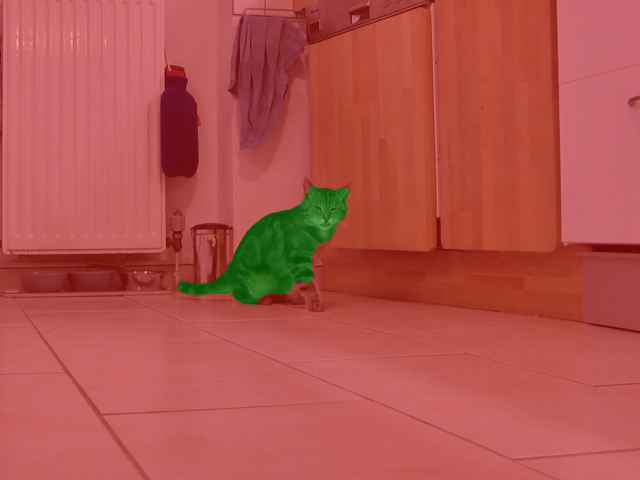}&
\includegraphics[height=1.9cm]{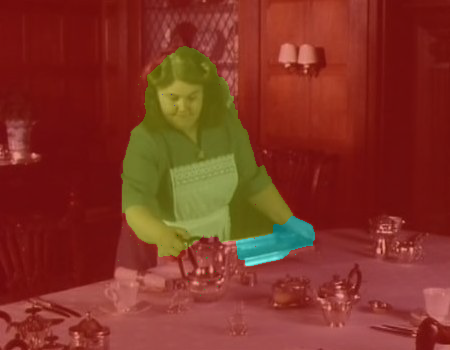}&
\includegraphics[height=1.9cm]{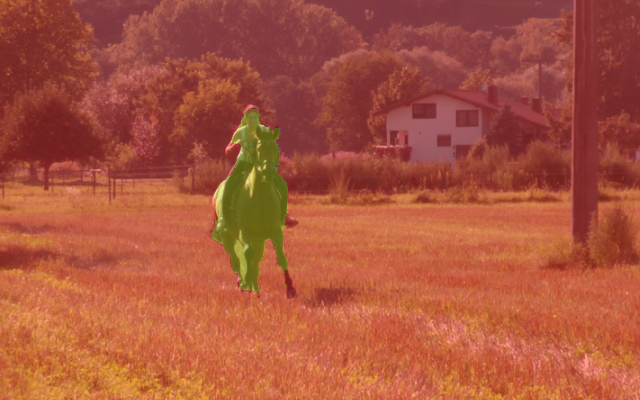}&
\includegraphics[height=1.9cm]{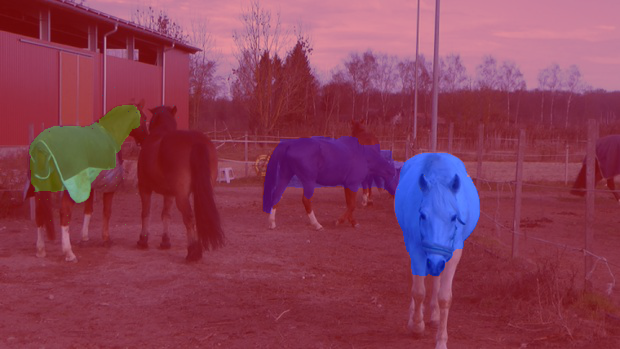}&
\includegraphics[height=1.9cm]{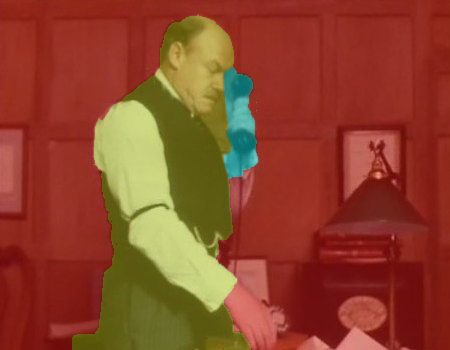}
\end{tabular}
}
\caption{Samples of our Lifted AOMC segmentation results densified by \cite{OB11}. Even for articulated motion, our segmentations show little over-segmentation.}
\label{fig:densified}
\end{figure*}
\begin{figure}[t]
 \centering
 \begin{tabular}{@{}c@{\,}c@{\,}c@{}}
 \includegraphics[width=0.33\linewidth]{imageshorses05_0070} &
 \includegraphics[width=0.33\linewidth]{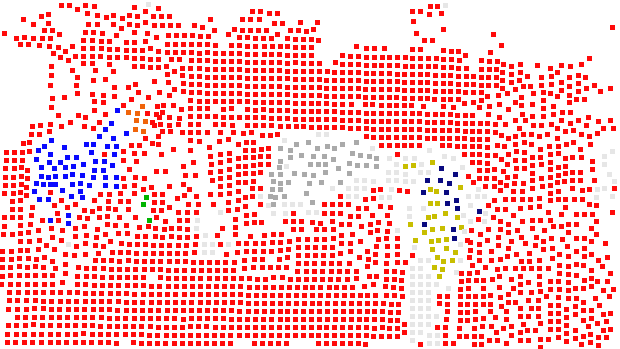}&
 \includegraphics[width=0.33\linewidth]{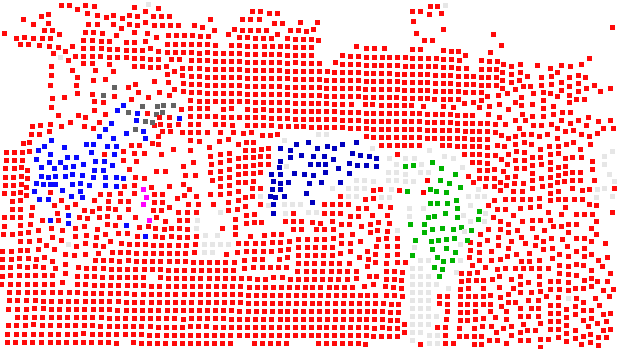}\\
 & MCe \cite{keuper15a}& ours
 \end{tabular}
 \caption{The scaling motion of the white horse moving towards the camera causes over-segmentation with a simple motion model~\cite{keuper15a}. 
 With the proposed Lifted AOMC, this can be avoided.}
 \label{fig:scalingm}
\end{figure}

Fig. \ref{fig:scalingm} gives an example of the segmentation quality under scaling. In the {\it horses05} sequence, scaling is caused by the motion of the white horse towards the camera. This causes over-segmentation in the competing method MCe \cite{keuper15a}, which can not handle higher-order motion models. With the proposed Lifted AOMC, the segmentation can be improved.

\begin{figure}[t]
\centering
\begin{tabular}{@{}c@{\,}c@{\,}c@{}}
\includegraphics[width=0.33\linewidth]{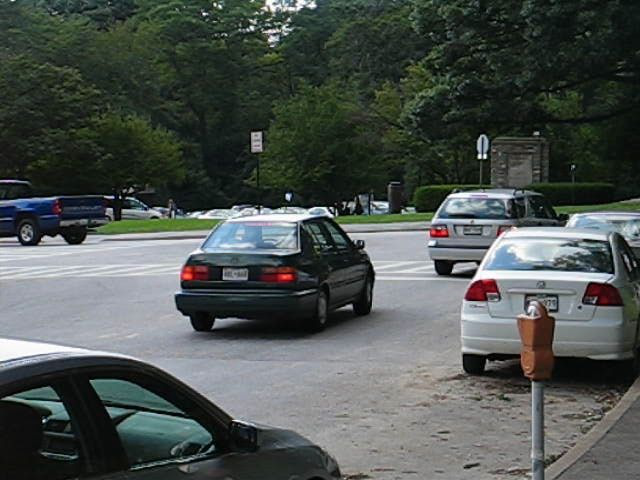} &
\includegraphics[width=0.33\linewidth]{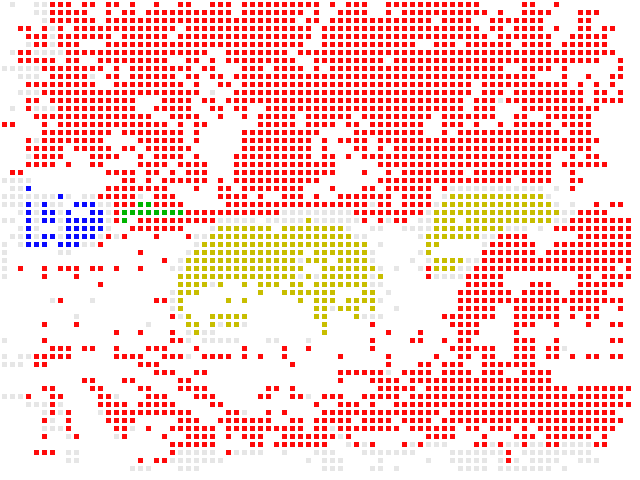}&
\includegraphics[width=0.33\linewidth]{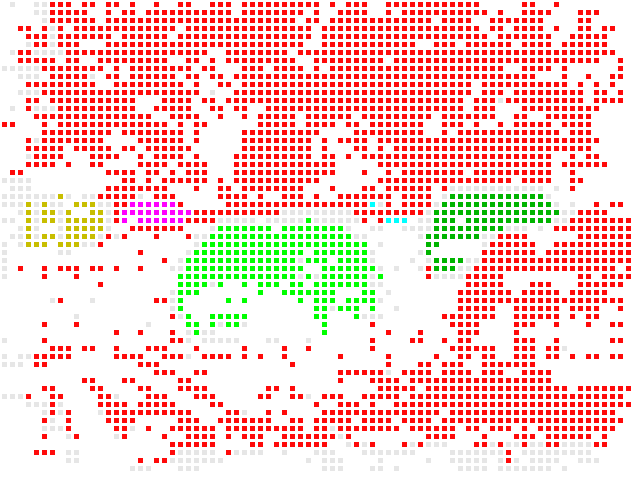}\\
& MCe \cite{keuper15a}& ours
\end{tabular}
\caption{The two cars in front move to the same direction, leading an assignment to the same cluster with the non-lifted mulitcut approach \cite{keuper15a}. 
The Lifted AOMC can assign the different cars to distinct segments.}
\label{fig:lifted}
\end{figure}

\begin{figure}[t]
\centering\begin{tabular}{@{}c@{\,}c@{\,}c@{}}
\includegraphics[width=0.33\linewidth]{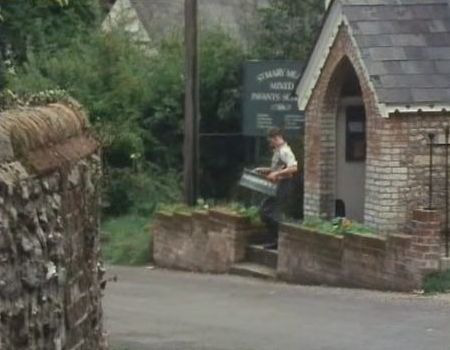} &
\includegraphics[width=0.33\linewidth]{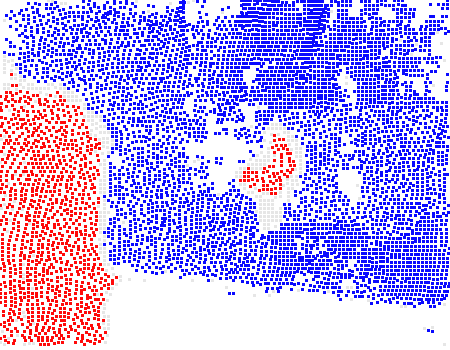}&
\includegraphics[width=0.33\linewidth]{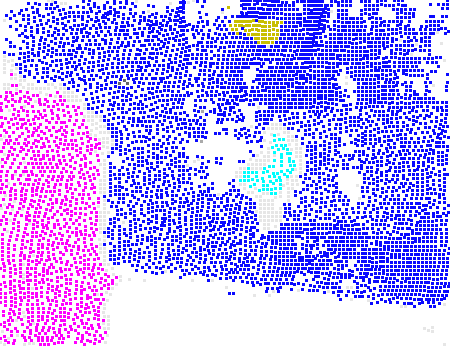}\\
& MCe \cite{keuper15a}& ours\end{tabular}
\caption{Due to camera motion the person and the wall are assigned to the same cluster with the non-lifted multicut approach \cite{keuper15a}. 
The Lifted AOMC allows for correct segmentation.}
\label{fig:lifted2}
\end{figure}

Fig. \ref{fig:lifted} and \ref{fig:lifted2} both show examples, where the same label is assigned to distinct objects that move similarly. In the {\it cars2} sequence in Fig. \ref{fig:lifted}, this is indeed due to similar real world object motion, whereas, in the {\it marple10} sequence, the effect is due to camera motion and the scene geometry. In both cases, the formulation of the Lifted AOMC problem allows to tell the distinct objects apart. However, in the {\it marple10} sequence (Fig.~\ref{fig:lifted2}), we can observe a spurious segment in the background, which is probably caused by unprecise flow estimation.  

\begin{figure}[t]
\centering\centering\begin{tabular}{@{}c@{\,}c@{\,}c@{}}
\includegraphics[width=0.33\linewidth]{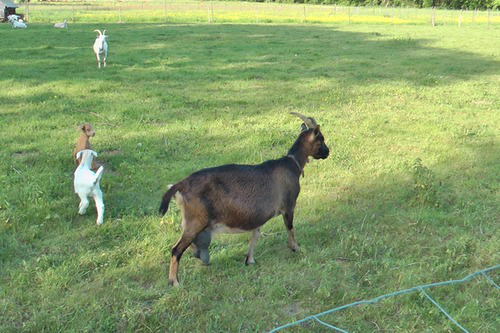}&
\includegraphics[width=0.33\linewidth]{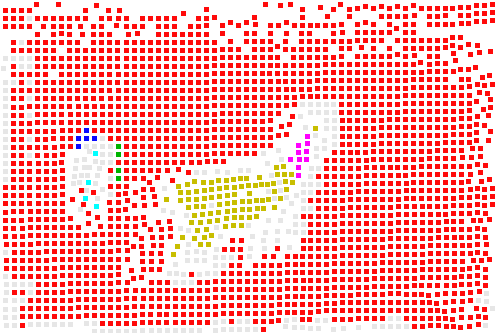}&
\includegraphics[width=0.33\linewidth]{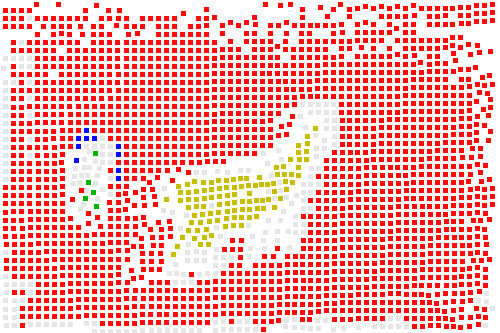}\\
& MCe \cite{keuper15a}& ours\end{tabular}
\caption{The articulated motion causes over-segmentation in \cite{keuper15a}. 
The Lifted AOMC performs better.}
\label{fig:ho2}
\end{figure}
\begin{figure}[t]
\centering
\centering\centering\begin{tabular}{@{}c@{\,}c@{\,}c@{}}
\includegraphics[width=0.33\linewidth]{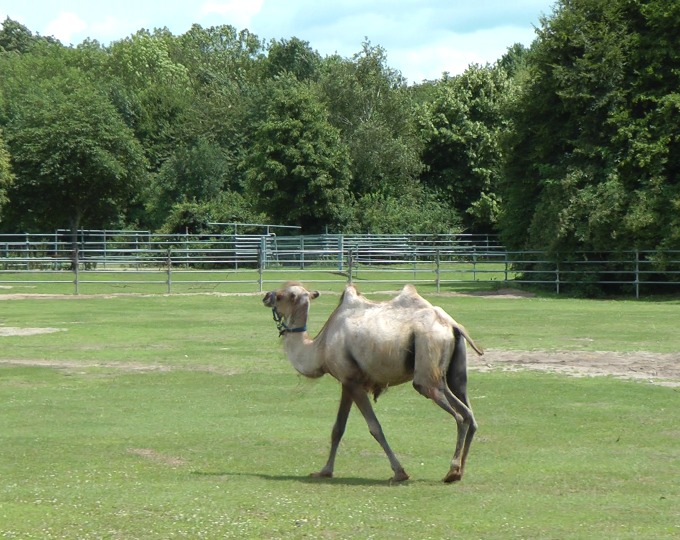} &
\includegraphics[width=0.33\linewidth]{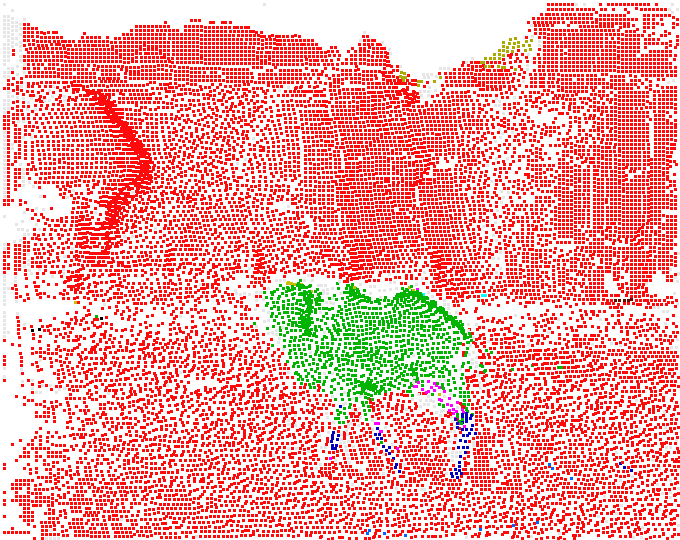}&
\includegraphics[width=0.33\linewidth]{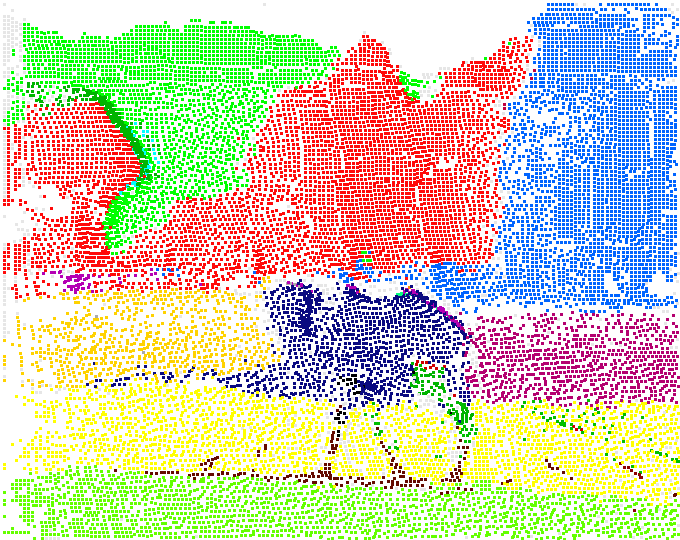}\\
& MCe \cite{keuper15a}& ours\end{tabular}
\caption{Failure case. The dominant camera motion causes strong over-segmentation with the proposed method. 
Here, our third order model can not model the  motion appropriately.}
\label{fig:failure}
\end{figure}

In Fig. \ref{fig:ho2}, we show an example of the {\it goats01} sequence. Here, the head and body of the goat in front are segmented into a distinct components by the pairwise method \cite{keuper15a}, because of the expressed articulated motion. Although our third order model can not explicitly handle articulation, the over-segmentation can be fixed in this case. 

In contrast, Fig. \ref{fig:failure} shows a failure case of the proposed method. Due to the dominant camera motion in a scene with complex geometry, the Euclidean motion model fits particularly badly. Thus our model leads to the segmentation of the scene into its depth layers, and thus to strong over-segmentation.

Several examples of pixel-segmentations computed from our sparse segmentation using \cite{OB11} are given in Fig. \ref{fig:densified}. The densified segmentations look reasonable. On the bear example, the articulated leg motion still causes some over-segmentation. One of the horses in the {\it horses05} sequence is missed. However, even small object such as the tray in the {\it marple12} sequence or the phone in the {\it marple13} sequence can be correctly segmented.

\paragraph{Results on Video Segmentation} 
To prove the benefits of our approach over pure pairwise terms as used in \cite{keuper15a}, we additionally compare both approaches on the DAVIS benchmark~\cite{Perazzi2016} validation set. Here we achieve
\begin{table}[tb]
\centering
\begin{tabular*}{\linewidth}{@{\extracolsep{\stretch{1}}}lcccc@{\extracolsep{\stretch{1}}}}
\toprule
&MCe~\cite{keuper15a}&    Ours\\
\midrule
Jmean &   0.552 &    0.565 \\
Jrecall  & 0.575    &  0.614\\
Jdecay   &0.022    & 0.020\\
Fmean  & 0.552    & 0.558\\
Frecall &  0.61   &    0.622\\
Fdecay  & 0.034  &  0.052\\
\bottomrule
\end{tabular*}
\caption{\label{tab:davis} Results on the DAVIS~\cite{Perazzi2016} video segmentation benchmark.}
\end{table}
Note that the task on DAVIS is to segment a single salient object, while both \cite{keuper15a} and our approach aim at segmenting all moving objects such that the results of both are below the state-of-the-art.

\subsection{Scalability}
In the following, we want to evaluate our proposed heuristic for the higher-order minimum cost lifted multicut problems (compare Alg.~\ref{alg:KL_outer}) on the FBMS-59 dataset~\cite{Ochs14} in terms of computation times. First, we report absolute computation times of our full pipeline of the lifted AOMC problem. Additionally, we generate toy problem instances for single frame motion segmentation of lifted HOMC problems for a more detailed analysis. 

\begin{figure}[t]
\centering
\includegraphics[width=\linewidth]{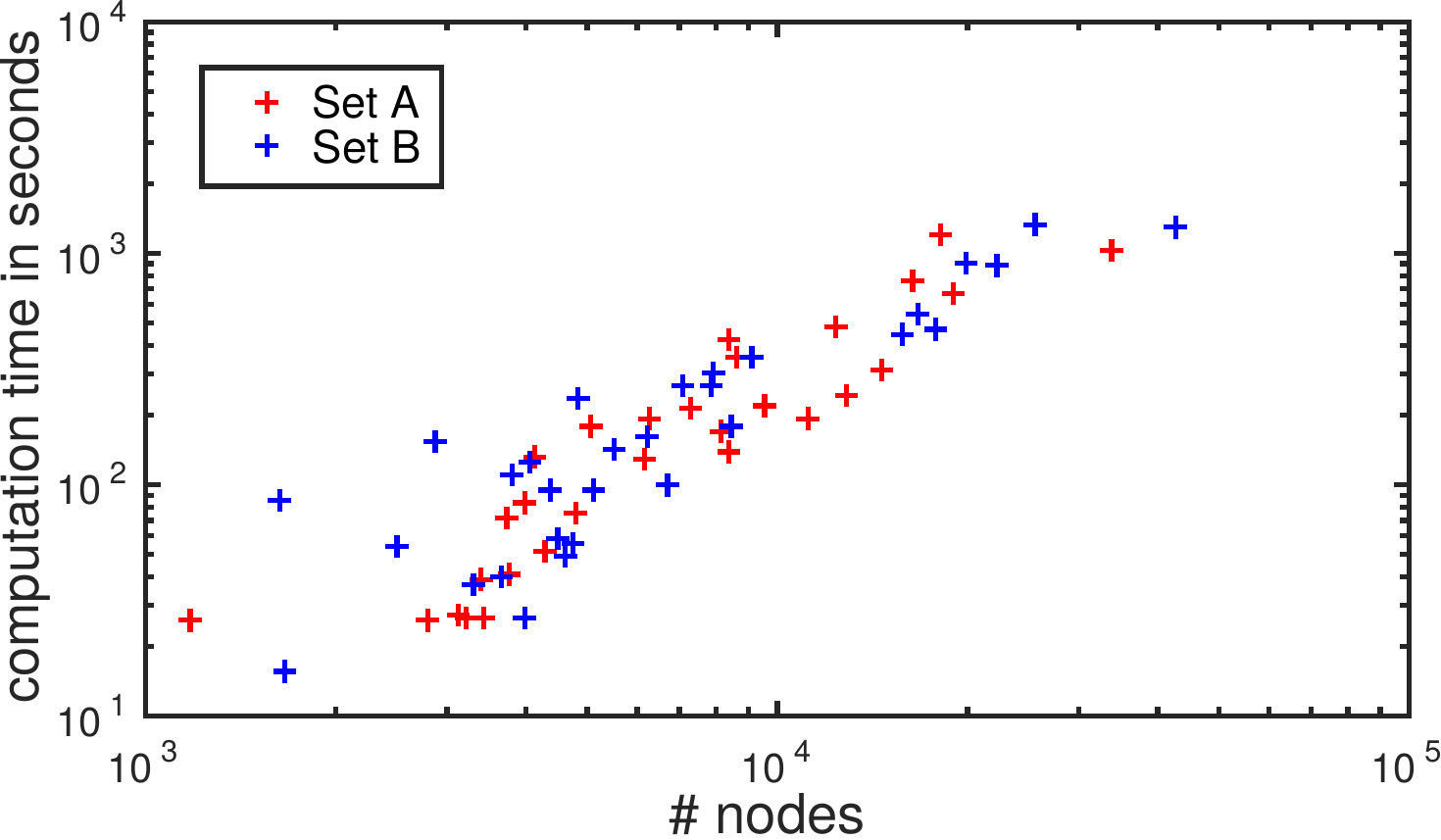}\caption{Computation times in logarithmic scale of the problem instances from Set A and Set B with respect to the number of nodes in logarithmic scale, \ie clustered trajectories. \label{fig:scaling}}
\end{figure}

\paragraph{Lifted AOMC}
Fig.~\ref{fig:scaling} shows a plot of the computation times of our full pipeline on Set A and B w.r.t. the number of clustered point trajectories in logarithmic scale. The runtime distribution indicates linear runtime behavior. However, the number of large problem instances is too small to make any claim. Yet, the plot shows that heuristic solutions can be generated in a few minutes for most instances.

\paragraph{Instance Computation}
\label{sec:instances}
\begin{figure}[t]
\centering
\begin{tabular}{@{}l@{}l@{}}
\includegraphics[width=0.48\linewidth]{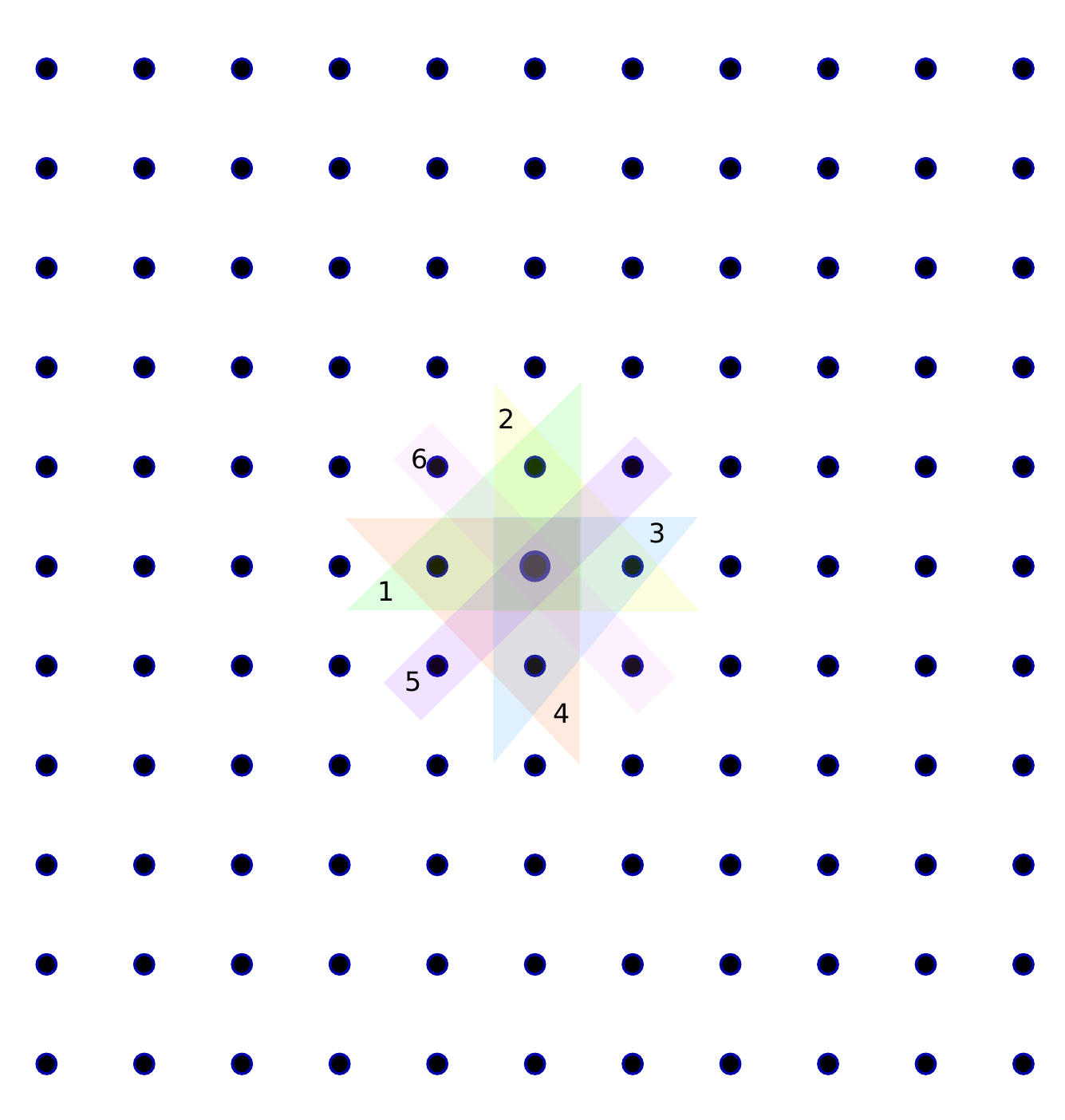}&
\includegraphics[width=0.48\linewidth]{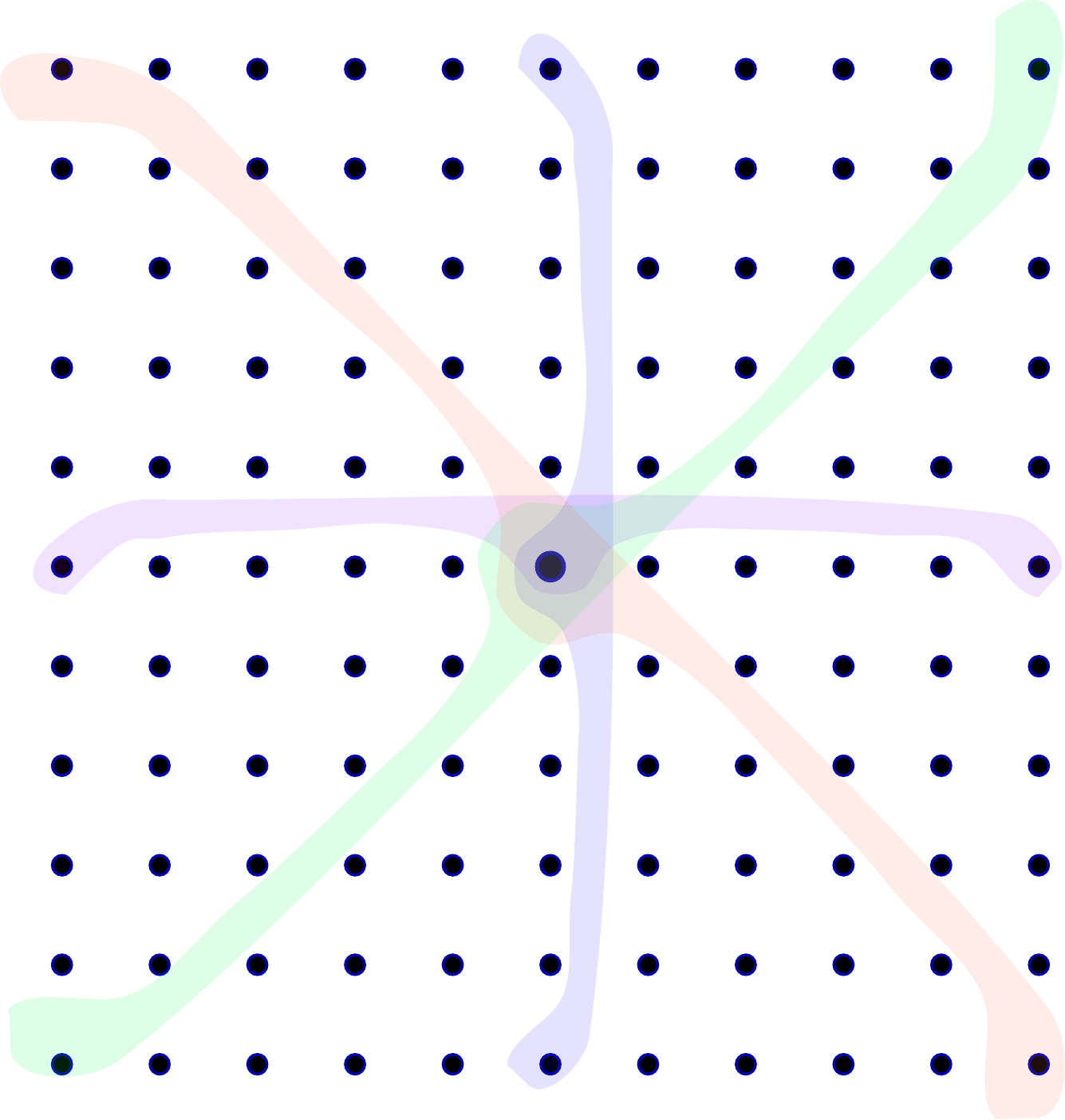}\\
(a) higher order edges&(b) lifted higher order edges
\end{tabular}
\caption{Higher order edges and lifted edges for the timing evaluation. \label{fig:edges}}
\end{figure}
\begin{figure}[t]
\centering
\begin{tabular}{@{}l@{}l@{}}
\includegraphics[width=0.48\linewidth]{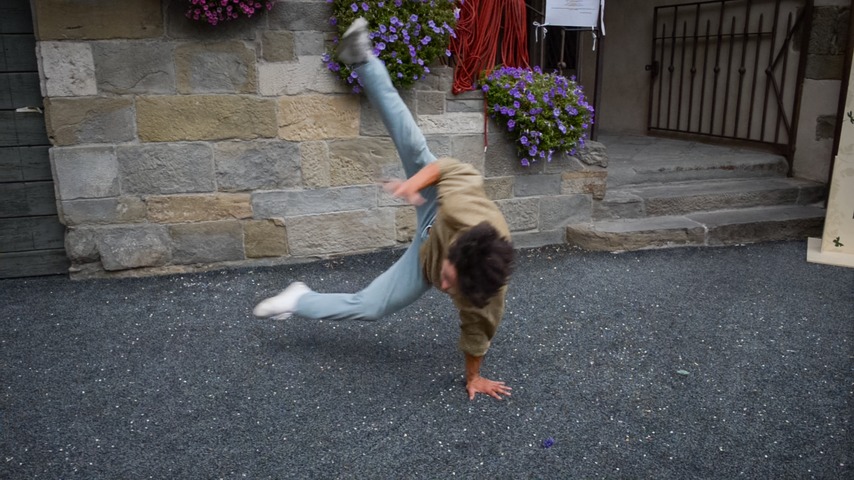}&
\includegraphics[width=0.48\linewidth]{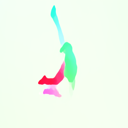}\\
image&optical flow\\
\includegraphics[width=0.48\linewidth]{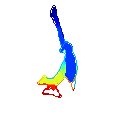}&
\includegraphics[width=0.48\linewidth]{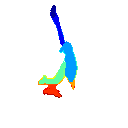}\\
result of non-lifted problem\hspace{0.3cm}&\hspace{0.3cm}result of lifted problem
\end{tabular}
\caption{Example from the \emph{breakdance-flare} sequence from the DAVIS \cite{Perazzi2016} dataset. From the successively optical flow, we generate non-lifted and lifted problem instances. Exemplary solutions are displayed in the bottom row. \label{fig:example}}
\end{figure}
From every of the 50 sequences of the DAVIS \cite{Perazzi2016}  dataset, we compute the optical flow between the first two frames using \cite{flownet}. To generate problem instances of varying sizes, we downsample the resulting flow fields to size $2^k\times2^k$ for $k=3,\dots,8$. 
Then, we build graphs $G = (V, F)$, where every pixel location is represented by a node in $V$. For every node, we insert 6 third order edges as depicted in figure \ref{fig:edges}~(a) to build $F$. Additionally, we insert pairwise edges with costs 0 between all nodes in an 8-pixel neighborhood. 
\begin{figure*}[t]
\centering
\begin{tabular}{@{}l@{}l@{}}
\includegraphics[width=0.48\linewidth]{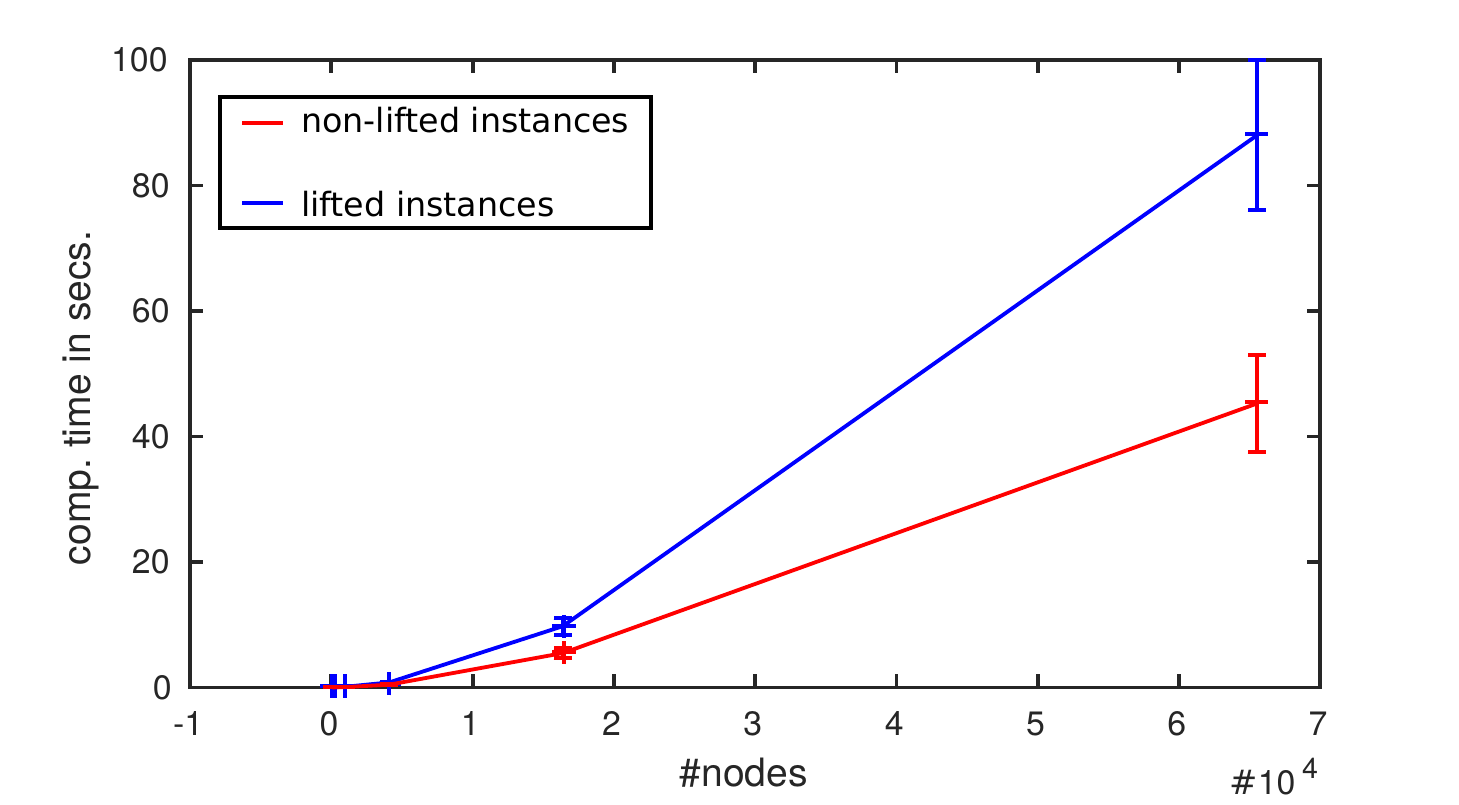}&
\includegraphics[width=0.48\linewidth]{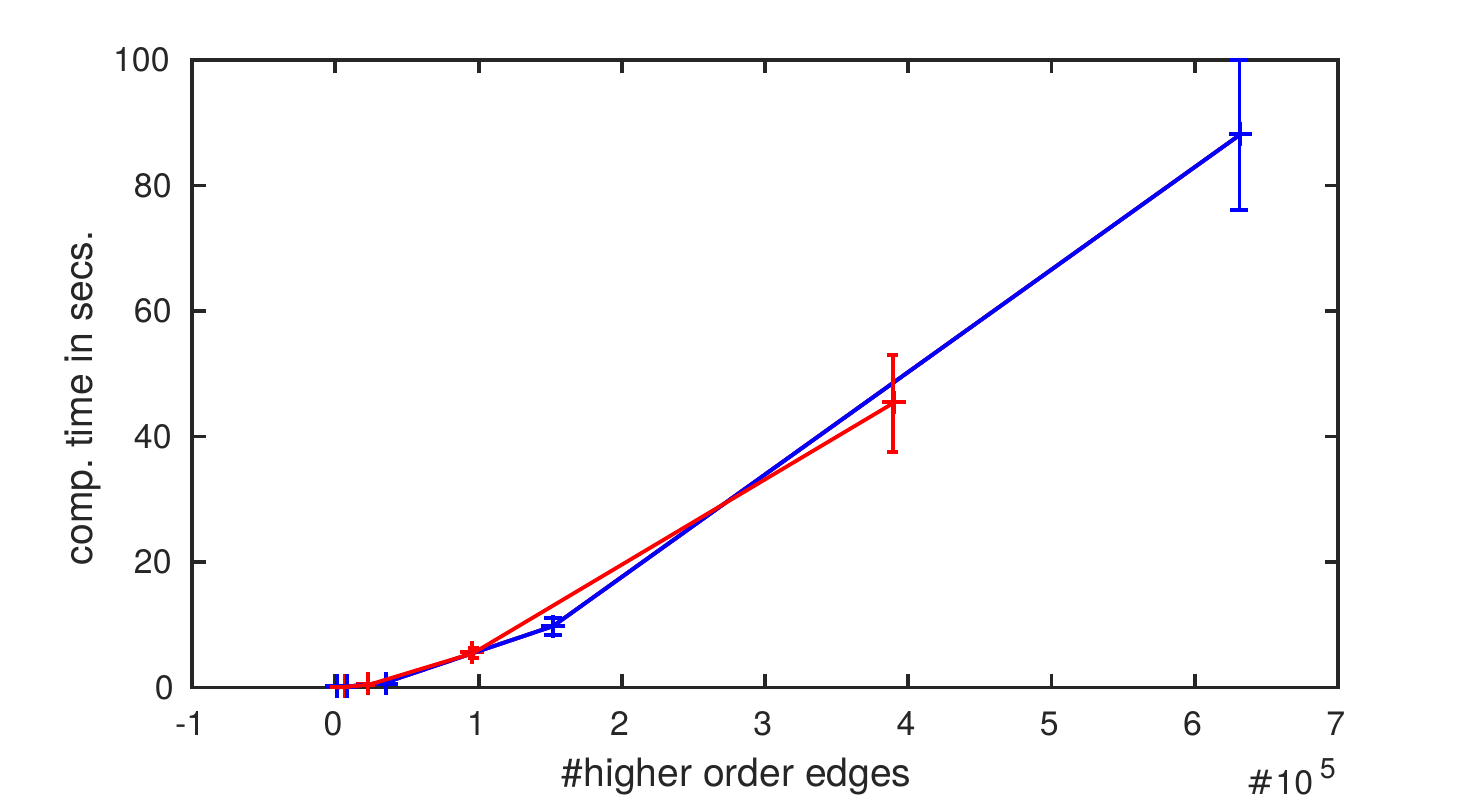}\\
(a)&(b)
\end{tabular}
\caption{Inference times in seconds over the number of nodes and the number of higher-order edges. Errorbars indicate the standard error. \label{fig:timing}}
\end{figure*}
We generate an additional set of lifted edges $F'$ by computing all horizontal and diagonal edges with exactly 5 pixel distance as depicted in figure  \ref{fig:edges}~(b). The resulting lifted graphs are denoted $G'=(V,F\cup F')$.

The costs for all third order edges are computed from optical flow vectors w.~r.~t. equation \eqref{eq:thirdordercosts}. 

An example image frame and the corresponding downsampled flow field to $128\times 128$ pixels in shown in Fig. \ref{fig:example}~(top). The clustering results from the proposed Alg.~\ref{alg:KL_outer} of the corresponding non-lifted problem (defined w.~r.~t. $G$) and the lifted problem (defined w.~r.~t. $G'$) are given in Fig. \ref{fig:example}~(bottom). As expected, the segmentation computed from the lifted problem instance looks very reasonable while the segmentation from the non-lifted problem is incorrect along the region boundaries.

\paragraph{Inference Times} 
\label{sec:timing}
For the 300 instances of non-lifted higher-order minimum cost multicut problems and the 300 lifted problem instances, we compute heuristic solutions with our generalization of the Kernighan-Lin method (Alg.~\ref{alg:KL_outer}) on a single Intel Xeon CPU with 2.70GHz. The solutions are initialized trivially by assigning all nodes to one clusters (all initial edge labels equal 1). The resulting computation times are given in \ref{fig:timing} w.~r.~t. the number of nodes (a) and w.~r.~t. the number of third order edges (b). The computation times of the lifted problem instances are significantly higher than the non-lifted problem instances when evaluated w.~r.~t. the number of nodes. However, when evaluated w.~r.~t. the total number of higher-order edges, the increase in compuation time is only linear. All instances have been solved in less than 150 seconds.

\section{Conclusion}
In this paper, we have presented a generalization of the minimum cost lifted multicut problem to a higher-order minimum cost lifted multicut problem, with an application to motion segmentation. 
For this new problem class, we have proposed a heuristic solver that allows to generate solutions on dense point trajectory graphs. 
Further, for the motion segmentation application, we have proposed an algorithm that allows to insert higher-order edges adaptively w.r.t. pairwise motion differences.
With this approach, we improve over the state-of-the-art in motion segmentation on the FBMS-59 benchmark. 
Since the proposed problem can model highly expressive cost functions, we hope that it is going to be useful for further computer vision applications.

{\small
\bibliographystyle{ieee}
\bibliography{paper}
}

\end{document}